\documentclass[letterpaper,10pt,conference]{ieeeconf}
\IEEEoverridecommandlockouts
\usepackage{fontawesome5}
\usepackage{graphicx}
\usepackage{times}
\usepackage{graphicx,xcolor}
\usepackage{multicol}
\usepackage{amsmath}
\usepackage{amssymb}
\usepackage{booktabs}
\usepackage{graphicx}
\usepackage{bbm}
\usepackage{multirow}
\usepackage{siunitx}
\usepackage{url}
\usepackage{capt-of}
\usepackage{caption}
\usepackage{siunitx}
\usepackage{adjustbox}
\usepackage[pagebackref=true,breaklinks=true,bookmarks=false]{hyperref}

\usepackage{algorithm}
\usepackage{algorithmicx}
\usepackage{algpseudocode}

\usepackage{bm}

\definecolor{pmcolor}{RGB}{70,70,70}

\newcommand{\paragraphc}[1]{\vspace{0.13em}\noindent\textbf{#1}}
\newcommand{\ourteleop}[0]{HATO}

\definecolor{FR}{HTML}{EDE9F5}
\definecolor{FE}{HTML}{844380}
\definecolor{S}{HTML}{B13D2D}
\definecolor{H}{HTML}{BADCFF}
\definecolor{FL}{HTML}{E48B50}

\newcommand{\secv}{\vspace{0.0em}}
\newcommand{\ssecv}{\vspace{0.0em}}

\title{\LARGE \bf
Learning Visuotactile Skills with Two Multifingered Hands
}
\author{
Toru Lin, Yu Zhang$^{*}$, Qiyang Li$^{*}$, Haozhi Qi$^{*}$, Brent Yi, Sergey Levine, and Jitendra Malik 
\thanks{* Equal contribution.}
\thanks{All authors are with University of California, Berkeley. Correspondence to {\tt toru@berkeley.edu}}%
}

\begin{document}

\let\oldtwocolumn\twocolumn
\renewcommand\twocolumn[1][]{%
\oldtwocolumn[{#1}{
\begin{center}
    \vspace{-1.5em}
    \includegraphics[width=\linewidth]{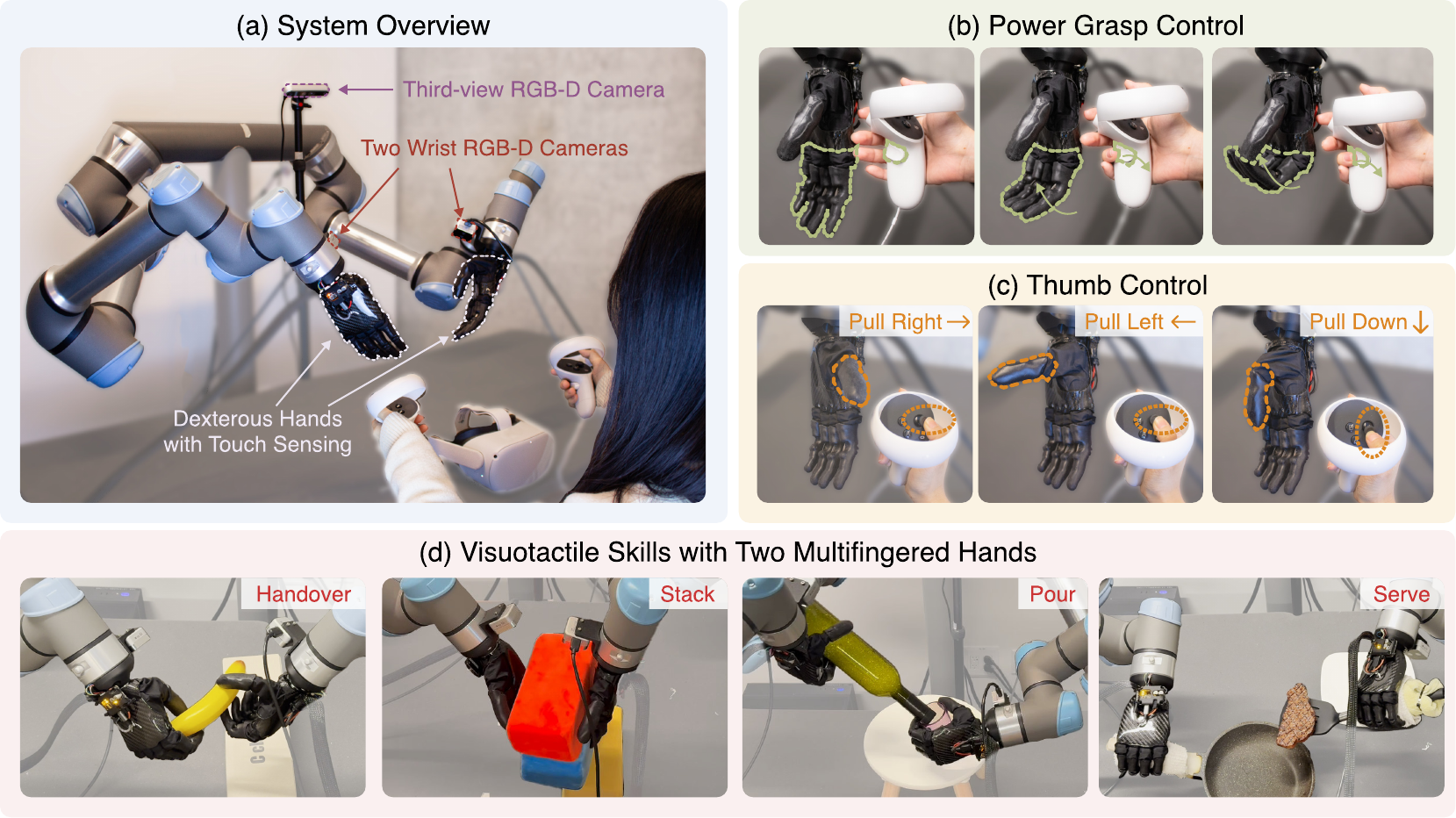}
    \captionof{figure}{\textbf{An overview of our system setup and learned visuotactile skills on four tasks.} (a) Our hardware and teleoperation system setup. The hardware consists of two UR5 robot arms, each equipped with a Psyonic Ability Hand. Visual observations are obtained via two wrist-mounted and one third-view RGB-D cameras. Tactile observations come from the multifingered hands, with each fingertip equipped with six touch sensors. We utilize the Meta Quest 2 platform for teleoperation. (b) We use the grip buttons of the Quest controllers to command power grasp of the non-thumb fingers. (c) We use thumbsticks to control the 2-DoF joint positions of the thumbs. (d) Four policies learned from visuotactile data collected by our hands-arms teleoperation system (HATO). These policies can accomplish a variety of complex bimanual tasks: handing over a slippery object, stacking a block tower, pouring from a wine bottle, and serving steak.}
   \label{fig:teaser}
   \vspace{-0.1em}
\end{center}
}]
}

\maketitle

\begin{abstract}
Aiming to replicate human-like dexterity, perceptual experiences, and motion patterns, we explore learning from human demonstrations using a bimanual system with multifingered hands and visuotactile data. 
Two significant challenges exist: the lack of an affordable and accessible teleoperation system suitable for a dual-arm setup with multifingered hands, and the scarcity of multifingered hand hardware equipped with touch sensing.
To tackle the first challenge, we develop \ourteleop{}, a low-cost \underline{h}ands-\underline{a}rms \underline{t}ele\underline{o}peration system that leverages off-the-shelf electronics, complemented with a software suite that enables efficient data collection; the comprehensive software suite also supports multimodal data processing, scalable policy learning, and smooth policy deployment.
To tackle the latter challenge, we introduce a novel hardware adaptation by repurposing two prosthetic hands equipped with touch sensors for research.
Using visuotactile data collected from our system, we learn skills to complete long-horizon, high-precision tasks which are difficult to achieve without multifingered dexterity and touch feedback. 
Furthermore, we empirically investigate the effects of dataset size, sensing modality, and visual input preprocessing on policy learning. 
Our results mark a promising step forward in bimanual multifingered manipulation from visuotactile data. Videos, code, and datasets can be found \href{https://toruowo.github.io/hato/}{here}.
\end{abstract}

\secv
\vspace{-0.6em}
\section{Introduction}
\secv

\begin{figure*}[t]
    \centering
    \includegraphics[width=\linewidth]{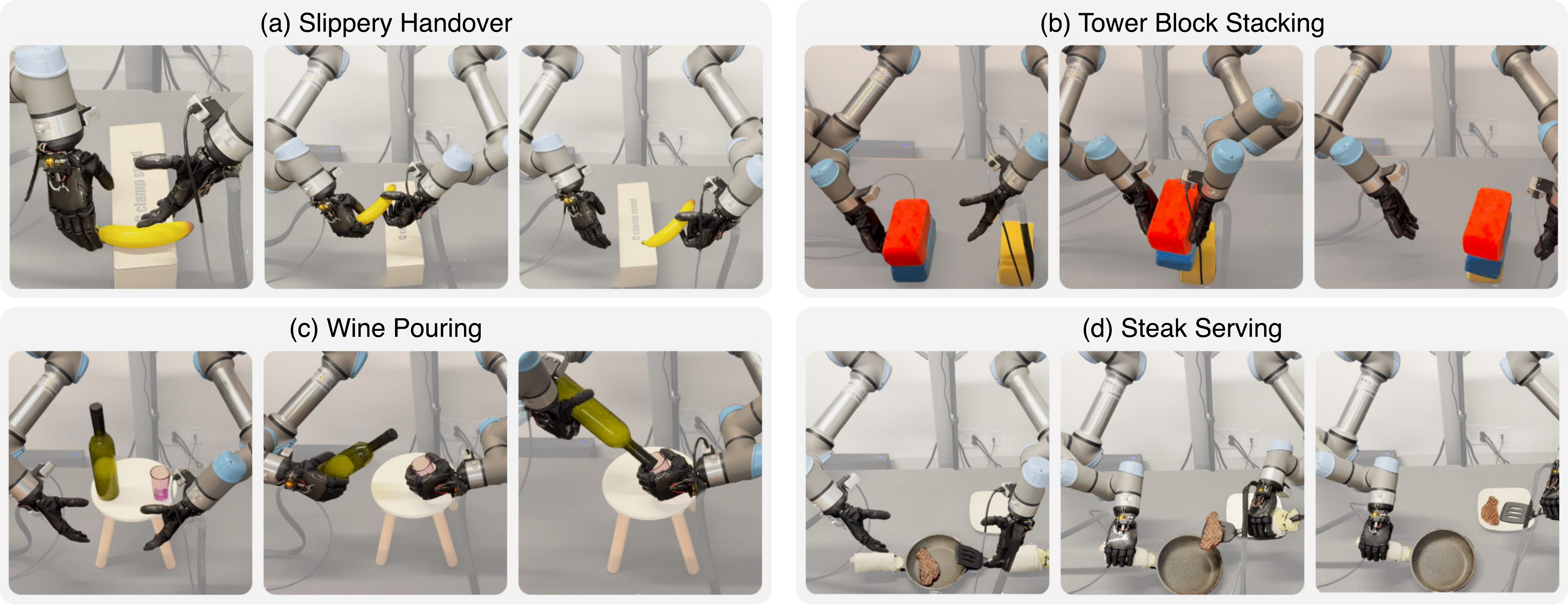}
    \caption{\textbf{Illustration of learned skills on four different tasks.} Our learned policies complete long-horizon and high-precision tasks that require bimanual dexterity. (a) \textit{Slippery Handover} tests the basic coordination skill between two hands. (b) \textit{Tower Block Stacking} demonstrates the advantage of having a large contact area because of the flat palm, as well as the ability to maintain a stable pose during movements. (c) \textit{Wine Pouring} requires the ability to grasp larger objects with the center of mass changing during manipulation. (d) \textit{Steak Serving} requires the ability to use tools and maintain grasp stability when moving the objects on the tool.}
    \label{fig:tasks}
    \vspace{-1em}
\end{figure*}

Achieving human-level dexterity is a long-term goal in the field of robotic manipulation. To this end, we explore the novel integration of a bimanual system with multifingered hands, visuotactile data modalities, and learning from human demonstrations, in hopes of mimicking the complexities of human anatomy, sensory experiences, and behavior patterns.

Most existing bimanual systems opt for parallel-jaw grippers~\cite{zhao2023learning,fu2024mobile,chi2024universal} as the end effectors, due to the high maintenance costs and limited availability of more advanced alternatives~\cite{allegro,schunk,shadow}. However, this choice greatly constrains the range of motions that can be achieved compared to multifingered hands~\cite{bicchi2000robotic}, limiting abilities in adaptive grasping, in-hand manipulation, dexterous handovers, tool use, and bilateral coordination for complex tasks.

We assemble a bimanual system with multifingered hands to learn dexterous skills through visuotactile demonstrations. We confront two major challenges: the absence of an affordable and accessible teleoperation system suitable for a dual-arm setup with multifingered hands, and the scarcity of multifingered hand hardware equipped with touch sensing. Our main contributions include: (1) a novel hardware adaptation that repurposes prosthetic hands equipped with rich tactile sensing for research use; (2) HATO, a low-cost \underline{h}ands-\underline{a}rms \underline{t}ele\underline{o}peration system built with commercially available virtual reality (VR) hardware, featuring novel mappings from teleoperator motion to robot control; (3) a comprehensive and versatile software suite for reliable and efficient data collection, multimodal data processing, scalable policy learning, and smooth policy deployment.

We design four complex tasks that examine our system's capability in achieving intricate skills like two-hand coordination, manipulation of bulky objects, and sophisticated tool use. From only 30 minutes to 2 hours of teleoperation data collected using \ourteleop{} (including around 5 to 10 minutes of practice time), we are able to obtain dexterous bimanual manipulation policies with visuotactile observations that can adeptly complete all tasks through pure end-to-end learning. Our system demonstrates natural and human-like skills and showcases unprecedented dexterity.

We also perform a thorough ablation study on the effects of dataset size, sensing modality, and visual input preprocessing on policy learning. Most notably, we find that vision and touch significantly enhance learning efficiency, policy success rate, and policy robustness. Without touch or vision, the policies are not able to consistently succeed or sometimes completely fail, highlighting the importance of high-quality touch sensing for enabling human-level dexterity. Our experiments also reveal that a dataset comprising a few hundred demonstrations is sufficient for learning effective bimanual dexterous policies. Additionally, we confirm the vital role of wrist-mounted cameras in boosting policy performance, while noting that depth information does not markedly benefit the learning process.

In the spirit of fostering further research and collaboration, we will open-source all our hardware and software systems and make the teleoperation dataset we have collected publicly available to facilitate evaluation and replication of our work.

\secv
\section{Related Work}
\secv

\paragraphc{Bimanual hands-arms manipulation.}
Prior bimanual manipulation systems usually use parallel-jaw grippers~\cite{zhao2023learning,Caccavale2008SixDOFIC,Sarkar1993DynamicCO,platt2004manipulation} for their simplicity and durability. Early work uses classical control for dexterous manipulation~\cite{ott2006humanoid,steffen2010bimanual,vahrenkamp2011bimanual}, but these approaches are largely task-specific and require expert knowledge of system dynamics and task structures. Reinforcement learning can alleviate these issues, but they tend to have high sample complexity and are only tractable in simulation~\cite{chen2022bidex,zakka2023robopianist}. Although there has been progress in transferring the policies from simulation to the real-world for a single hand~\cite{qi2022hand,chen2022visual,openai2018learning,handa2023dextreme} or two hands~\cite{huang2023dynamic,lin2024twisting}, the policies usually suffer from the sim-to-real gap.

\paragraphc{Imitation learning.} Another promising approach to achieve human-level dexterity is learning from demonstrations~\cite{billard2008survey,hussein2017imitation}. More recently, researchers have utilized deep neural networks to obtain better representations and policies~\cite{ravichandar2020recent}. While bimanual manipulation has also been studied in this context~\cite{zhao2023learning,fu2024mobile,chi2024universal,grannen2023stabilize}, these studies only demonstrate systems with parallel-jaw grippers. Wang et al.~\cite{wang2024dexcap} is concurrent work that showcases bimanual dexterous manipulation skills. Unlike \cite{wang2024dexcap}, which utilizes only vision and proprioception as inputs, our work additionally utilizes tactile input and shows that touch sensing is critical for the reliable completion of many challenging tasks considered.

\paragraphc{Bimanual teleoperation.} Having access to a diverse set of high-quality demonstrations is crucial for learning a high-quality policy. Existing teleoperation systems are largely restricted to the use of parallel-jaw grippers~\cite{zhao2023learning, fu2024mobile, fang2023low, seo2023deep,wu2023gello,arunachalam2023dexterous} or single-handed scenarios~\cite{arunachalam2023holo, qin2022one, sivakumar2022robotic,iyer2024open}. In addition, the hand teleoperation systems heavily rely on retargeting to solve morphology differences, thus introducing unavoidable latency and are not intuitive to use. In contrast, our system teleoperates a robot hand by separating finger control into thumb control and power grasp control, providing a smoother user experience.

\paragraphc{Learning visuotactile skills.} Our system also features rich visuotactile sensory data. Integrated visuotactile sensing has been used for numerous applications such as grasping~\cite{Calandra2018More}, in-hand manipulation~\cite{qi2023general}, object shape reconstruction~\cite{smith2021active,suresh2023neural,suresh2022shapemap} and object recognition~\cite{xu2023tandem3d}, cloth manipulation~\cite{sunil2023visuotactile}, and learning representation for object interactions~\cite{guzey2023dexterity,guzey2023see}. However, none of them use two multifingered hands, and previous bimanual manipulation systems are either not equipped with dexterous hands or lack rich sensory signals. To the best of our knowledge, our work is the first at the intersection of bimanual dexterous manipulation and imitation learning from visuotactile inputs.

\begin{figure}[!t]
    \centering
    \includegraphics[width=\linewidth]{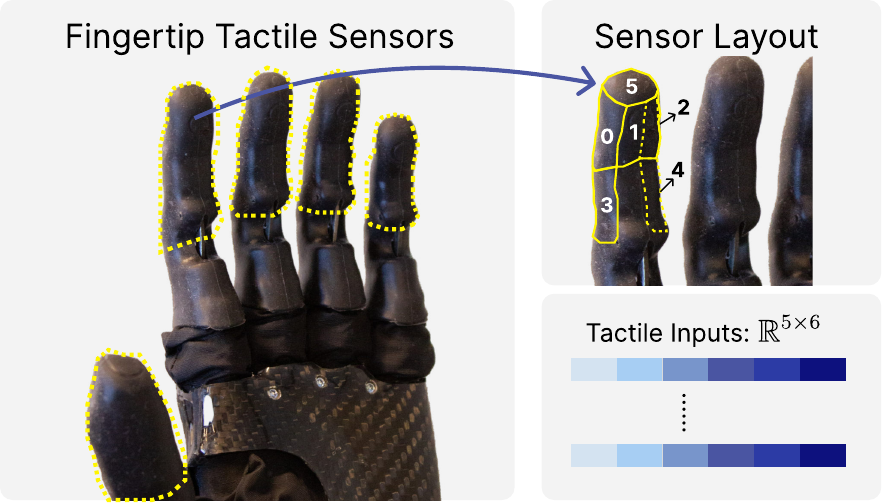}
    \caption{\textbf{Fingertip Tactile Sensor Layout.} There are six tactile sensors on each of the fingertips. Each tactile sensor provides a continuous value proportional to the sensed pressure.}
    \label{fig:touch}
    \vspace{-1em}
\end{figure}

\secv
\section{HATO: \underline{H}ands-\underline{A}rms \underline{T}ele-\underline{O}peration}
\secv

\begin{figure*}[!t]
    \centering
    \includegraphics[width=\linewidth]{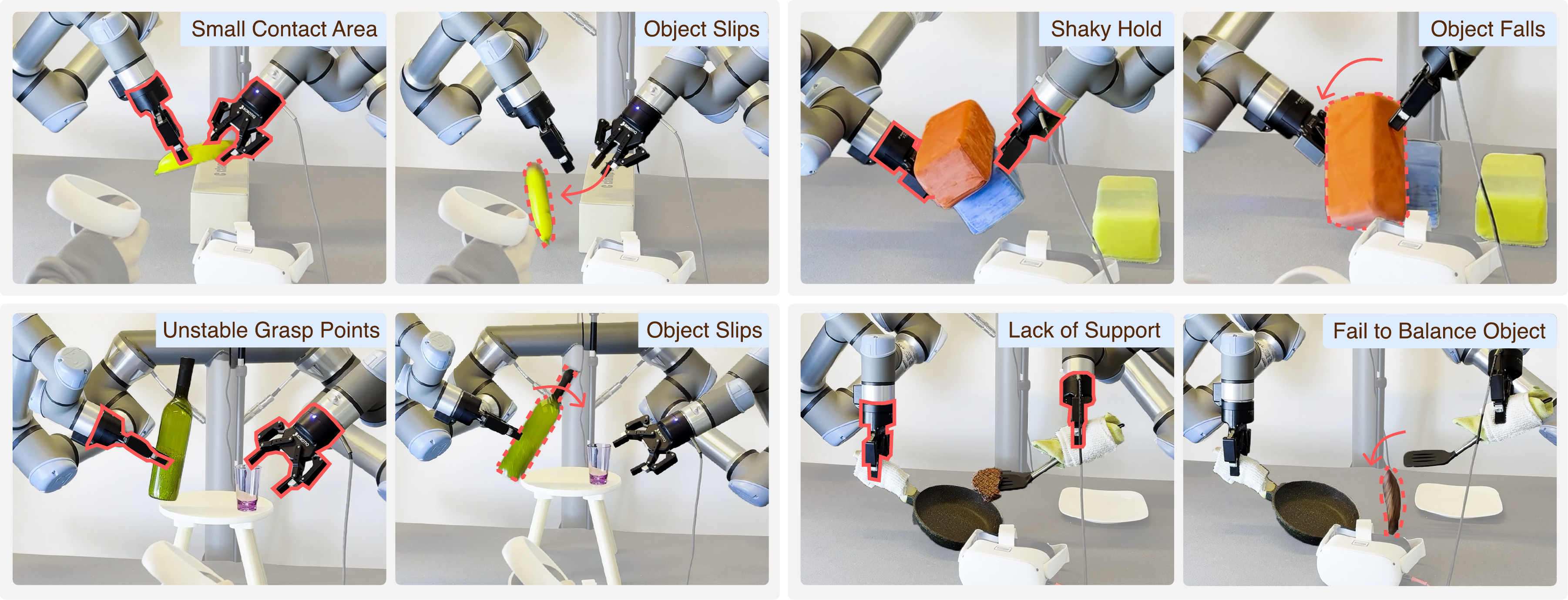}
    \caption{\textbf{Common failures from parallel-jaw gripper teleoperation.} When the object has a slippery and rigid surface or is larger than the gripper, grasping with a parallel-jaw gripper requires very accurate planning, which is difficult to achieve during teleoperation. This difficulty leads to various failure modes, including slipping objects, unstable holds, and unstable grasps. In contrast, multifingered hands provide additional redundancy and contact areas to maintain a grasp.}
    \label{fig:gripper}
    \vspace{-1em}
\end{figure*}

We develop HATO, a novel teleoperation system for bimanual multifingered hands. Our system is easy to set up and intuitive to use, enabling efficient collection of bimanual dexterous manipulation data. An overview of our system is shown in Figure~\ref{fig:teaser}.
For teleoperation of each hand-arm pair, \ourteleop{} maps a Meta Quest 2 virtual reality (VR) controller's pose to the end-effector pose of the robot arm, and the controller's grip button and thumbstick to the hand's joint positions. The \ourteleop{} software suite includes a data collection pipeline that records and processes data from all available sensing modalities (vision, touch, and proprioception).
We provide details below on the hardware setup, teleoperation pipeline, and data infrastructure.

\ssecv
\subsection{Robot Setup}
\ssecv

\paragraphc{Robot arms.} We use two UR5e robot arms for our manipulation system. The UR5e is an industrial arm with six degrees of freedom (DoF). Each DoF is a revolute joint that has a working range of $[-2\pi, 2\pi]$.

\paragraphc{Robot hands.} We attach two Psyonic Ability Hands to the UR5e arms as end effectors. These hands were originally designed for prosthetic use~\cite{akhtar2021ability}; we repurpose them for research by designing custom printed circuit boards (PCBs) that simplify electrical wiring by integrating communication interfaces with power distribution. Each hand has five fingers, and each finger has six actuated DoFs (one DoF per finger, two for the thumb). Each actuated DoF of the hand is a revolute joint that has upper and lower limits similar to human finger limits (see the leftmost and rightmost columns of Figure~\ref{fig:teaser} (b)). For the four non-thumb fingers, the metacarpophalangeal (MCP) joints serve as the actuating DoF; the MCP joints connect to the proximal interphalangeal (PIP) joints via a four-bar linkage mechanism, contributing to an additional underactuated DoF on each finger. Each fingertip also comes with six touch sensors (see Figure~\ref{fig:touch}).

\ssecv
\subsection{Teleoperation Setup}
\label{sec:setup-teleop}
\ssecv
Our teleoperation system leverages the Meta Quest 2 platform. It comes with a VR headset and a pair of controllers, each designated for one hand. The Quest combines visual tracking with Inertial Measurement Unit (IMU) sensors to capture the spatial orientation and movements of the controllers. Each controller is also outfitted with a range of interactive buttons, including thumbsticks, trigger buttons, and grip buttons. Using a VR application like \textit{oculus\_reader}~\cite{oculus_reader}, one can stream data related to the controllers' poses and button states in real-time. Our main contribution is the development of a software suite that provides flexible options for translating movements detected by the Quest controllers to precise control commands for a bimanual multifingered robotic system. We highlight a simple yet effective mapping from controller buttons to multifingered hand movements, which enables a smooth and intuitive user experience.

\paragraphc{Arm control.} We first read the pose measurements from the Quest controller, then transform the pose to a desired end-effector (EEF) pose of the robot's coordinate system. We use inverse kinematics (IK) to solve for the joint positions given the desired EEF pose, and send the joint position command to the UR5e arm. In the case of IK solving failure, we use the last commanded joint positions. Aside from this control implementation, our software suite also contains two other implementations. The second implementation uses the first-order approximation of the IK solution which linearly maps the delta pose between the desired EEF pose and the current EEF pose of the arm to the delta positions of the joints. The third implementation directly sends the end-effector position to the arm and the IK is done onboard. For the rest of the paper, we use the first implementation for both teleoperation and policy execution.

\paragraphc{Hand control.} We map the controller's grip button to the joint positions of the four non-thumb fingers (4 DoF), and map the thumbstick readings to joint positions of the thumb (2 DoF). Readings from the grip button and thumbstick of each controller are re-normalized based on the joint ranges of their corresponding hand DoF, and sent to the Ability Hand as position control targets. Specifically, pressing/releasing the grip button controls flexion/extension of the non-thumb fingers (Figure \ref{fig:teaser} (b)), and the 2-D positions of the thumbstick control the thumb joints' flexion/extension and abduction/adduction (Figure \ref{fig:teaser} (c)). The hand's power grasp is therefore a continuous movement proportional to the operator's pressing force on the grip button, allowing fine-grained control when grasping soft or deformable objects. While such mapping sacrifices the ability to perform sophisticated finger-gaiting, it provides an intuitive user interface and is still able to complete various grasps for complex tasks~\cite{feix2015grasp}.

\paragraphc{Pause-and-adjust.} We follow a conventional design where we use the controller's trigger button to start and break a continuous arm control sequence, allowing the teleoperator to pause during a teleoperation trial and adjust their posture. This design greatly helps with tasks that are close to the teleoperator's physical limit, e.g., those that require a large reach of the teleoperator's arm.

Such mapping also makes \ourteleop{} flexible enough to control potentially any robot arms with EEF control and hands with an independent anthropomorphic thumb and the ability to perform power grasps.

\ssecv
\subsection{Data Collection and Preprocessing for Policy Learning}
\ssecv

We collect multimodal data from both hands and arms by running \ourteleop{} data collection pipeline at 10Hz. The data include the proprioceptive states of both the UR5e arms and the Ability Hands, the RGB-D images from three RealSense depth cameras (two mounted on the wrist of each hand, one mounted at a stationary ``head-view'' position), the touch sensor readings from the Ability Hands, and the control commands given to the UR5e arms and the Ability Hands.

\paragraphc{Proprioception.} Our proprioceptive state data at each time step includes the current joint positions of both arms, the current finger positions of the hands, and the current end-effector pose (represented as a concatenation of translation and axis-angle).

\paragraphc{Vision.} We obtain RGB-D image data from three cameras at each time step; both the RGB and depth images are streamed at a resolution of $480 \times 640$. We resize all images to $240 \times 320$ before feeding them into the network.

\begin{figure*}[!t]
    \centering
    \includegraphics[width=\linewidth]{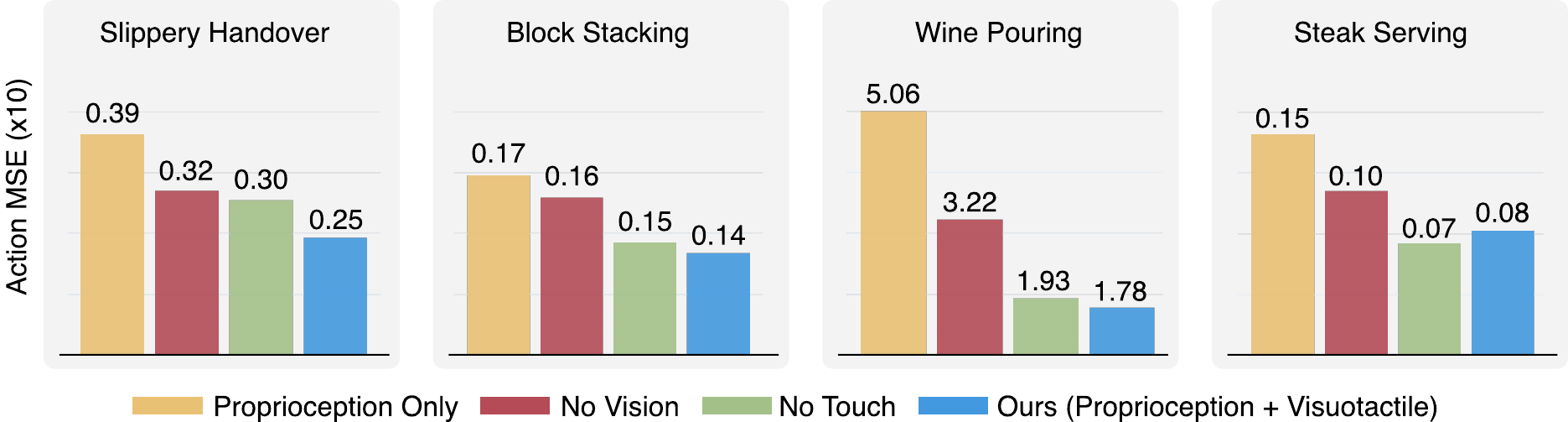}
    \caption{\textbf{How vision and touch affect the policy performance across four challenging tasks.} Across all tasks, vision is crucial for the policy to achieve low prediction error. For \textit{Block Stacking} and \textit{Steak Serving}, removing the touch input does not significantly influence the prediction error, but as we show in Table \ref{tab:init_block} and \ref{tab:vistac_stack}, touch is also crucial for successfully completing these two tasks.
    }
    \label{fig:abl_modailty}
    \vspace{-1.2em}
\end{figure*}

\paragraphc{Touch.} Each finger of the Ability Hand has six touch sensors attached (Figure~\ref{fig:touch}), contributing to a total of sixty touch sensor readings from both hands $h \in \mathbb{R}^{60}$. Each touch sensor reading is a continuous number whose range changes depending on manufacturing tolerances. The readings typically lie in the range of $[200, 400]$ (ADC values) when there is no contact event, and above $1000$ on contact.

\paragraphc{Control actions.} For each arm, the control action is the desired joint position for each of the six revolute joints of the arm, where each value in the vector represents the angle of the revolute joint (in $[-2\pi, 2\pi]$). For each hand, the control action is the desired joint position for each of the six finger joints (independent of the arm control implementation).

\paragraphc{Data normalization.} All values are scaled linearly per dimension to be between $-1$ and $1$, with the minimum value mapped to $-1$ and the maximum value mapped to $1$. The minimum and maximum values are obtained from the training data, except for the joint position reading and the control action of the hand. We use a minimum value of 0 and maximum values of $[110, 110, 110, 110, 90, 120]$ for each of the six finger joint positions.
For depth and RGB images, we use their raw values: $[0, 255]$ for the RGB images and $[0, 65535]$ for the depth images.

\secv
\section{Learning Visuotactile Skills with \ourteleop{}}
\secv

With visuotactile demonstration data collected from \ourteleop{}, we can learn a variety of bimanual dexterous skills for complex tasks. In this section, we describe how we train diffusion policies~\cite{chi2023diffusion} using the collected data. Importantly, we also propose a novel asynchronous inference algorithm, which is the key to our fast and smooth policy deployment.

\ssecv
\subsection{Learning}
\ssecv

Our system learns bimanual dexterous skills from demonstration data by treating action prediction as a conditional generation problem. Following \cite{chi2023diffusion}, with an observation horizon of $1$, we predict the action sequence of length $16$ from the current input observations using a denoising diffusion probabilistic model (DDPM)~\cite{ho2020denoising}. Each input observation is a collection of observation data from multiple modalities: proprioception, vision, and touch. Each action is a $24$-dimensional vector that specifies the desired joint positions for the two arms and the two hands. We choose to use only a single observation input as opposed to a short horizon of observations (as done in \cite{chi2023diffusion}) as we have found that using one single observation is sufficient for the policy to perform well and is much faster to train.

\paragraphc{Proprioception.}
We use end-effector poses as the proprioception observations and do not include the arm joint positions. This is because the UR5 arms do not have redundant joints, and the joint positions can vary very unpredictably near singularity during teleoperation, posing bigger learning challenges. The proprioception is passed through a two-layer network with ReLU activation, a hidden size of $256$, and an output feature size of $64$.

\paragraphc{Touch.} The touch signal is also passed through a two-layer network in the same way as proprioception.

\paragraphc{Vision.} For image and depth observations from the three cameras, we follow the prior work on diffusion policy~\cite{chi2023diffusion} to use the ResNet-18 architecture~\cite{he2016deep} and replace all the BatchNorm~\cite{ioffe2015batch} in the network with GroupNorm~\cite{wu2018group}. The fully connected layer's output size is adjusted to be $32$. We do not share network weights across camera inputs.

\paragraphc{Diffusion architecture.} All the encoded image, depth, tactile, and proprioceptive observations are then concatenated as the input to a diffusion model (CNN-based in \cite{chi2023diffusion}). We also use the same noise schedule (square cosine schedule) and the same number of diffusion steps (100) for training.

\paragraphc{Output action.} The diffusion output from the model is the normalized 6 DoF absolute desired joint positions of each UR5e arm, and the 6 DoF normalized ($0$ to $1$) desired joint positions of each Ability hand.

\paragraphc{Optimization details.} We use the AdamW optimizer~\cite{kingma2014adam, loshchilov2017decoupled} with a learning rate of $0.0001$, weight decay of $0.00001$, and a batch size of 128. Following \cite{chi2023diffusion}, we maintain an exponential weighted average of the model weights and use it during evaluation/deployment.

\ssecv
\subsection{Deployment}
\ssecv
At deployment time, we use an asynchronous setup where the diffusion model prediction and the robot execution run in parallel. In particular, we use a remote inference server that keeps track of the most recent observation and the timestep to which the observation corresponds. The local process sends the new observation to the remote inference server at every control step. The inference server continuously runs the diffusion model on the latest observation and produces the action sequence prediction. Then, it sends the action sequence prediction (with the corresponding timesteps) to the local process, where it computes the average of the predictions over multiple timesteps (similar to the temporal ensemble in \cite{zhao2023learning}). Note that this is different from how deployment is done in \cite{chi2023diffusion}, where they do not use a temporal ensemble. We have found that the inclusion of action aggregation greatly improves motion smoothness. For inference, we use 15 diffusion steps.

\secv
\section{Experiments}
\secv

We consider four challenging real-world tasks (Figure~\ref{fig:tasks}) to study the bimanual dexterity enabled by our system. We validate the effectiveness of our system setup, data collection pipeline, policy learning, and deployment pipelines by demonstrating teleoperation capabilities and showcasing learned skills that successfully complete these tasks. We also conduct an empirical investigation on how policy performance is influenced by data size and sensing modalities.

\paragraphc{\emph{Slippery Handover.}} Handing over objects is a motor skill commonly required for a wide range of daily activities; it is also commonly used as a bimanual manipulation task~\cite{li2023efficient}.
Each task episode is initialized with the slippery object resting on a box. One hand needs to pick up the object and hand it over to the other hand. The episode succeeds when the other hand holds the object stably and moves away from the first hand by a distance of more than $10$ centimeters.
As we will demonstrate, parallel grippers can face more challenges when handling objects with slippery surfaces compared to multifingered hands.
The anthropomorphic hand morphology greatly mitigates the slippery challenge due to the larger contact area and additional support that it provides to objects.

\begin{table}[!t]
\centering
\setlength{\tabcolsep}{7pt}
\renewcommand{\arraystretch}{1.3}
\resizebox{\linewidth}{!}{%
\begin{tabular}{rrrrr}
\toprule
{Task} & {Handover} & {Stacking} & {Pouring} & {Serving} \\
\cmidrule(r){1-1}
\cmidrule(l){2-5}
Pickup & 10 / 10 & 10 / 10 & 10 / 10 & 10 / 10 \\
Task Success & 10 / 10 & 10 / 10 & 9 / 10 & 5 / 10 \\
\bottomrule
\end{tabular}
}
\caption{\textbf{Success rate on each of the four challenging bimanual manipulation tasks.} For \textit{Slippery Handover} and \textit{Wine Pouring}, we use only image observation and proprioceptive state as we find these two inputs are sufficient to achieve an almost 100\% success rate. For \textit{Block Stacking} and \textit{Steak Serving}, we use image, proprioception, and touch as inputs. The pickup success is an intermediate metric that measures how often the hands successfully pick up both objects.}
\label{fig:main}
\vspace{-1em}
\end{table}

\paragraphc{\emph{Tower Block Stacking}.} Manipulation of bulky objects is another common skill required for many everyday tasks. Motivated by manual labor jobs in construction sites where workers need to use two hands to move bricks, we design a tower block stacking task. Each task episode begins with two piles of large blocks on the table, one consisting of two blocks (red and blue) and the other a single yellow block. The robot needs to move up the pile of two blocks and stack it on top of the yellow block. A successful episode is marked when the two moved blocks stay stably on the yellow block after being released from the robot hands.

\paragraphc{\emph{Wine Pouring}.} At task initialization, a large wine bottle and a small cup are placed on a stool on top of a table. The robot needs to use one hand to grasp the wine bottle, use the other hand to grasp the cup, perform a pouring motion from the bottle to the cup, and put both the bottle and the cup back. The wine bottle is filled with transparent beads to simulate liquid. Since the center of mass quickly changes during the pouring process, we hypothesize that a power grasp with a multifingered hand can greatly reduce the difficulty.

\paragraphc{\emph{Steak Serving}.} Inspired by cooking activities, we design a long-horizon task that requires prehensile grasps and intricate force feedback control loops for humans. The goal of this task is to serve a piece of cooked steak onto a plate. At task initialization, three objects are placed on the table: a cooking pan with a piece of steak inside, a spatula for serving, and a ceramic plate. The robot is expected to use one hand to hold the pan and the other hand to grasp the spatula. It then needs to insert the spatula under the bottom of the steak and lift it up. The task is completed when the robot successfully holds the spatula with the steak and serves it onto the plate.

\paragraphc{Teleoperation data collection details.} For \emph{Slippery Handover} and \emph{Tower Block Stacking}, we collect 100 demonstrations, each demonstration lasting around 6 seconds and 20 seconds respectively. For \emph{Wine Pouring} and \emph{Steak Serving}, we collect 300 demonstrations, each demonstration lasting around 25 seconds and 40 seconds respectively. Before the data collection for each task, we ask the human teleoperator to practice for 5 to 10 minutes.

\begin{figure}[t]
    \centering
    \includegraphics[width=\linewidth]{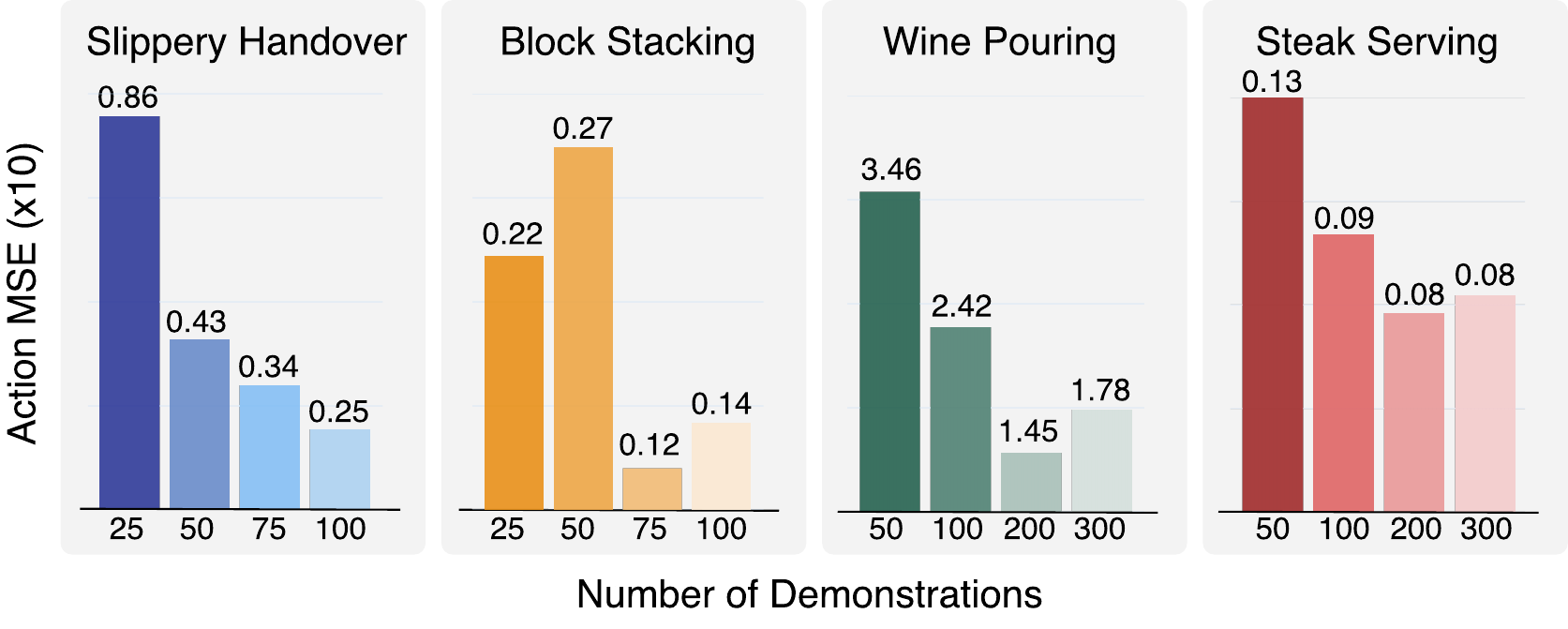}
    \caption{\textbf{How does the demonstration dataset size affect policy prediction error?} Across all tasks, having more demonstration trajectories consistently leads to lower prediction loss. In particular, the policy performance saturates for \emph{Block Stacking} at 75 demonstrations, \emph{Wine Pouring} at 200 demonstrations, and \emph{Steak Serving} at 100 demonstrations.}
    \label{fig:samples}
    \vspace{-1em}
\end{figure}

\begin{table*}[!t]
\adjustbox{valign=t}{
\begin{minipage}{.31\linewidth}
\centering
\setlength{\tabcolsep}{0.8em}
\renewcommand{\arraystretch}{1.2}
\resizebox{0.98\linewidth}{!}{%
\begin{tabular}{lcc}
\toprule
{Modality} & {Default Init.} & {Rare Init.} \\
\cmidrule(r){1-1}
\cmidrule(l){2-3}
Ours & 10 / 10 & 10 / 10 \\
w.o. Touch & 10 / 10 & 4 / 10 \\
w.o. Vision & 10 / 10 & 0 / 10\\
\bottomrule
\end{tabular}
}
\caption{\textbf{Success rate on \textit{Block Stacking} task.} We use two task initialization schemes (as illustrated in Figure~\ref{fig:init_block_viz}). The default initialization (\emph{Default Init.}) is more frequently encountered in the demonstration dataset compared to the other initialization (\emph{Rare Init.}). Without visual or touch feedback, the policy struggles to succeed from \emph{Rare Init.}}
\label{tab:init_block}
\end{minipage}
}
\hspace{0.75em}
\adjustbox{valign=t}{
\begin{minipage}{.31\linewidth}
\centering
\setlength{\tabcolsep}{1.3em}
\renewcommand{\arraystretch}{1.2}
\resizebox{0.98\linewidth}{!}{%
\begin{tabular}{lcc}
\toprule
{Modality} & {Pickup} & {Success} \\
\cmidrule(r){1-1}
\cmidrule(l){2-3}
Ours & 10 / 10 & 5 / 10 \\
w.o. Touch & 10 / 10 & 0 / 10 \\
w.o. Vision & 0 / 10 & 0 / 10 \\
EEF Only & 0 / 10 & 0 / 10 \\
\bottomrule
\end{tabular}
}
\caption{\textbf{Success rate on \textit{Steak Serving} task using policies trained with different sensing modalities.} The pickup success is an intermediate metric that measures how often the hands successfully pick up both objects. The policy cannot finish the task without all three sensing modalities and cannot pick up the object without touch.}
\label{tab:vistac_stack}
\end{minipage}
}
\hspace{0.75em}
\adjustbox{valign=t}{
\begin{minipage}{.31\linewidth}
\centering
\setlength{\tabcolsep}{1.3em}
\renewcommand{\arraystretch}{1.3}
\resizebox{0.98\linewidth}{!}{%
\begin{tabular}{lcc}
\toprule
{Modality} & {Pickup} & {Success} \\
\cmidrule(r){1-1}
\cmidrule(l){2-3}
Ours & 10 / 10 & 10 / 10 \\
w. Depth & 10 / 10 & 1 / 10 \\
Only Wrist & 0 / 10 & 1 / 10 \\
Only 3rd View & 0 / 10 & 0 / 10 \\
\bottomrule
\end{tabular}
}
\caption{\textbf{Success rate on \textit{Steak Serving} task with different camera configurations.} The pickup success is an intermediate metric that measures how often the hands successfully pickup both objects. We find having depth does not improve the performance and having all the cameras are necessary to finish this challenging task.}
\label{tab:camera_steak}
\end{minipage}
}
\vspace{-1em}
\end{table*}

\subsection{Capabilities from Teleoperation}

We qualitatively investigate whether having multifingered hands as end-effectors allows for better manipulation capabilities than parallel-jaw grippers by comparing their performances on the four manipulation tasks above. In particular, to test the manipulation capability of the parallel-jaw gripper, we keep the rest of the system the same while replacing the Ability hand with the Robotiq gripper and mapping the same grip button on the Quest controller to the gripper's open/close control. With multifingered hand end effectors, previously inexperienced teleoperators are able to collect hundreds of high-quality demonstrations within a few hours. On the other hand, with parallel-jaw grippers, the teleoperation suffers from a variety of failure modes, including slipping objects, shaky holds, unstable grasping, and unstable balancing. Some common task failures are shown in Figure~\ref{fig:gripper}.

\ssecv
\subsection{Capabilities from Learning}
\ssecv

We validate the effectiveness of \ourteleop{} as a data collection pipeline by demonstrating successful policies trained from \ourteleop{}-collected datasets. In particular, we record the task success rate of learned policies using 10 deployment trials. In addition to the success rate for the full task, we also record how many times each policy successfully picks up the object(s) (e.g., bottle and cup for pouring, pan and spatula for steak serving, two blocks for stacking, and banana for handover) as the partial task completion rate. As shown in Table~\ref{fig:main}, our policy is able to pick up the object(s) with a 100\% success rate across all tasks. Our policy is also able to complete three of the four tasks (handover, stacking, and pouring) consistently with near 100\% success rate. The last task, steak serving, is much more difficult to learn due to its long task horizon and the demand for high-precision control (e.g., balancing the steak on a spatula held by a hand). Despite such difficulty, our policy is still able to achieve around a 50\% success rate.
\ssecv
\subsection{Learning Efficiency}
\ssecv
We study the efficiency of our learning method by empirically evaluating the correlation between the number of demonstrations and policy performance. We use the mean squared error between predicted actions and ground truth actions (ActionMSE) on a held-out test set as the evaluation metric. The evaluation dataset consists of $10$ demonstration trajectories for \textit{Slippery Handover} and \textit{Block Stacking}, and $20$ demonstration trajectories for \textit{Wine Pouring} and \textit{Steak Serving}. The results are shown in Figure~\ref{fig:samples}. As we expect, the prediction error decreases as the training dataset size increases. It is worth highlighting that the ActionMSE metric roughly saturates for \textit{Block Stacking}, \textit{Steak Serving}, and \textit{Wine Pouring} at around 75, 100, and 200 demonstrations each, respectively. For the other tasks, \textit{Slippery Handover}, it is possible that more demonstrations may further improve policy robustness.

\ssecv
\subsection{Importance of Vision and Touch}
\ssecv
Our policy takes three types of sensing modalities as inputs: proprioception, vision, and touch. In this section, we quantitatively investigate how the visuotactile sensing modalities affect policy learning and performance.

\paragraphc{Vision and touch are crucial for learning.}
Figure~\ref{fig:abl_modailty} shows the ActionMSE for policies trained with different sensing modalities. For all four tasks, the policy trained with no vision has a much higher prediction error than the policy trained with all three sensing modalities. For all tasks except for \textit{Steak Serving}, the policy trained with no touch has a higher prediction error than the policy trained with all three sensing modalities. Policies trained with neither vision nor touch consistently have the highest ActionMSE. As we will show in the following section, such a correlation also translates to the policy success rate.

\paragraphc{Vision and touch improve policy success rate.}
On the \textit{Steak Serving} task, we evaluate the success rate of the policies trained with different sensing modalities and report the results in Table~\ref{tab:vistac_stack}. Without vision, the policy fails at the first task stage, i.e., properly picking up the objects (0/10 success rate). Without touch, the policy is able to accomplish the first task stage (i.e., pick up the pan and the spatula) but fails to transfer the steak over to the plate. It is worth highlighting that even though the ActionMSE metric is similar for the policies trained with or without touch ($0.07$ vs $0.08$), these policies have vastly different success rates: 0/10 (without touch) vs. 5/10 (with touch). We believe that this is because ActionMSE cannot fully capture how well the diffusion policy fits the dataset distribution. A potentially better metric would be to estimate the log-likelihood of the data under the diffusion policy output distribution, but that has been notoriously difficult to estimate~\cite{song2021maximum}. In our experience, we found that the ActionMSE was able to mostly inform us which policy is expected to perform well.

\begin{figure}[t]
    \centering
    \includegraphics[width=\linewidth]{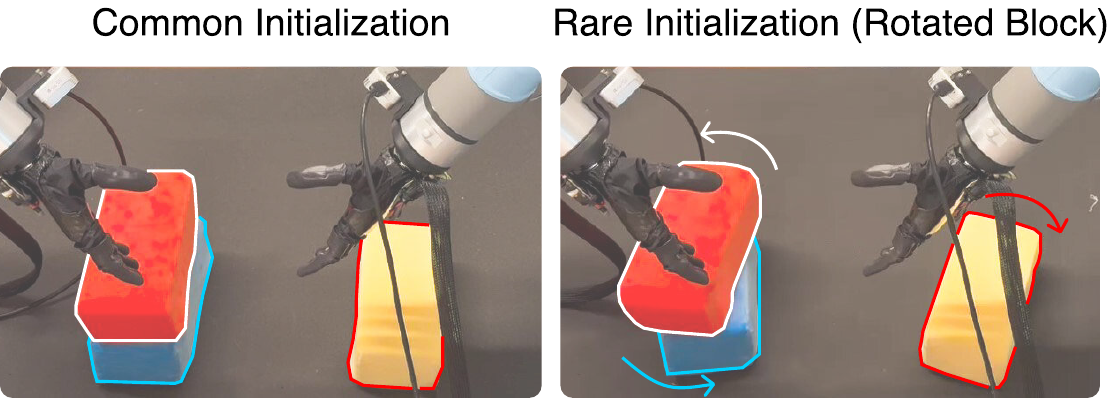}
    \caption{\textbf{Two scene initializations for the \texttt{Block Stacking} task.} The \emph{default initialization} is shown on the left and the \emph{rare initialization} is shown on the right. In the \emph{rare initialization}, the perturbed block configuration makes the two-block pile harder to pick up by two robot hands and more difficult to stack stably on another block.}
    \label{fig:init_block_viz}
\end{figure}

\paragraphc{Vision and touch improve policy robustness.}
To further understand how the sensing modality affects the robustness of our policy, we experiment with a less common scene initialization for the \textit{Block Stacking}. Specifically, we rotate the blocks in random directions. Figure~\ref{fig:init_block_viz} shows a comparison between the default initialization and the rare initialization. This scene initialization is less encountered in the demonstration dataset, and the perturbed block configuration makes the two-block pile harder to be picked up by two robot hands. In Table~\ref{tab:init_block}, we show a comparison of the success rate for this rare initialization and the default initialization across three different sensing configurations. While touch and vision are not needed for the default scene initialization (they all achieve 100\% success rate), for the rare scene initialization, the policy trained without touch can only succeed 4/10 times, and the policy trained without vision cannot succeed at all. This suggests that vision and touch sensing modalities allow the policy to be more robust to rare scenarios.

\begin{figure}[!t]
    \centering
    \includegraphics[width=\linewidth]{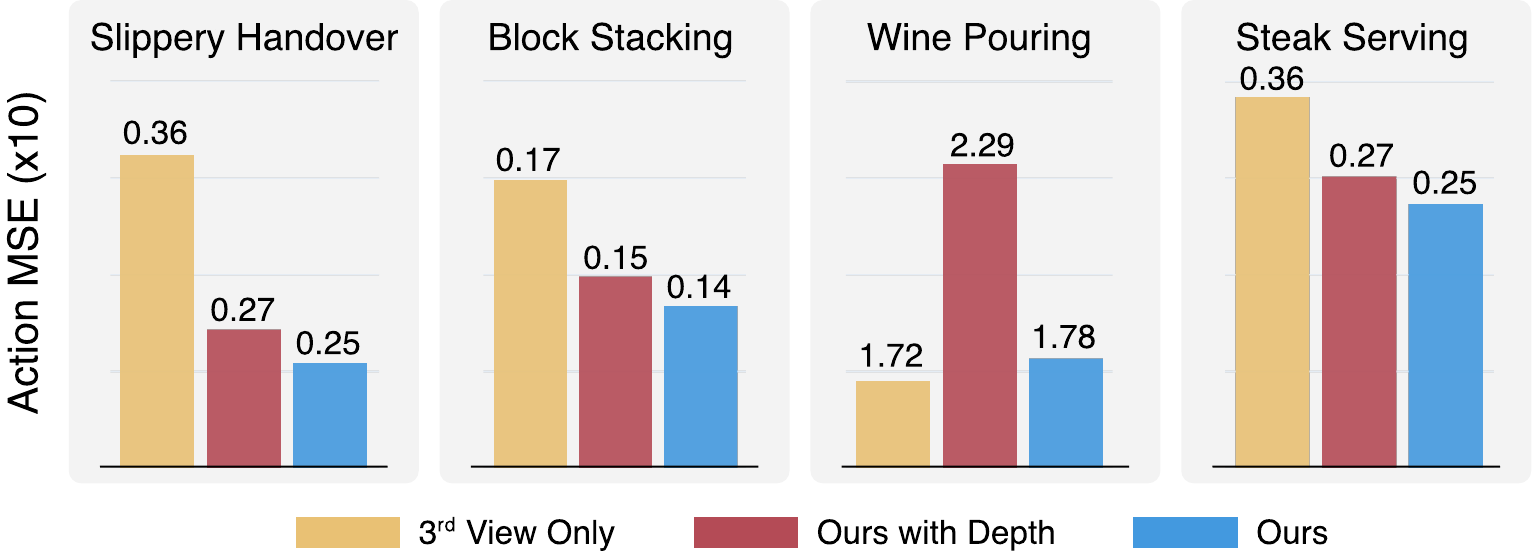}
    \caption{\textbf{How does each type of image observation affect policy prediction error?} Aside from the vision modality, all policies are trained with proprioception and touch. The policies trained with only the third-view camera (\emph{3$^{\text{rd}}$ View Only}) have higher prediction errors, except for the \emph{Wine Pouring} task. Policies with all three cameras that additionally include depth images (\emph{Ours with Depth}) have a similar prediction error to the policies trained without depth (\emph{Ours}), except on \emph{Wine Pouring}.}
    \label{fig:input-cam}
    \vspace{-1em}
\end{figure}

\paragraphc{Wrist camera vs third-view camera.} We study the effect of different camera positions. The results are shown in Figure~\ref{fig:input-cam}. Prediction error with only the wrist-view camera is consistently lower than that with only the third-view camera across the \textit{Slippery Handover}, \textit{Block Stacking}, and \textit{Steak Serving} tasks; the two errors are comparable on the \textit{Wine Pouring} task. We hypothesize that this is because wrist-view cameras contain richer information on task-relevant object states, due to the less occluded object view and more spatial hints via induced perspectives during arm movement.

\paragraphc{Use of depth.} We examine the prediction errors of the policies trained with and without depth information. The results are shown in Figure~\ref{fig:input-cam}. Across all four tasks, adding depth does not provide a marked benefit in terms of the ActionMSE metric, sometimes even hurting performance (e.g., \textit{Wine Pouring}). We hypothesize that the noisy depth readings cause more harm than good for learning.

\secv
\section{Discussions}
\secv

In this work, we share novel engineering insights that enable high-performance policy learning from human demonstrations using a bimanual system with multifingered hands and visuotactile sensing, and showcase the dexterous manipulation capabilities achieved by our system. In particular, we show that visuotactile sensing is the key for our policies to complete complex and long-horizon tasks consistently and robustly. Our low-cost teleoperation system opens up a number of avenues for future research. For example, equipping our teleoperation system with haptic feedback (e.g., attached to tactile sensing) could greatly improve the user experience and consequently the quality of teleoperation data. Furthermore, our policy is learned entirely from scratch with no pre-training, making it susceptible to appearance changes in the scene. Training more robust and generalizable policies is also an interesting future research direction.

\section*{Acknowledgment}

We thank Jesse Cornman from PSYONIC for help with setting up the Ability Hands, and Philipp Wu for help with setting up the UR5e robot arms and GELLO.
TL is supported by fellowships from the National Science Foundation and UC Berkeley.
QL is supported by ONR under N00014-20-1-2383, and NSF IIS-2150826.
HQ is supported by the DARPA Machine Common Sense and ONR MURI N00014-21-1-2801. This research was also partly supported by Savio computational cluster provided by the Berkeley Research Compute program.

\vspace{-0.3em}
\secv
\bibliographystyle{IEEEtran}
\bibliography{IEEEabrv,references}

\begin{thebibliography}{10}
\providecommand{\url}[1]{#1}
\csname url@rmstyle\endcsname
\providecommand{\newblock}{\relax}
\providecommand{\bibinfo}[2]{#2}
\providecommand\BIBentrySTDinterwordspacing{\spaceskip=0pt\relax}
\providecommand\BIBentryALTinterwordstretchfactor{4}
\providecommand\BIBentryALTinterwordspacing{\spaceskip=\fontdimen2\font plus
\BIBentryALTinterwordstretchfactor\fontdimen3\font minus \fontdimen4\font\relax}
\providecommand\BIBforeignlanguage[2]{{%
\expandafter\ifx\csname l@#1\endcsname\relax
\typeout{** WARNING: IEEEtran.bst: No hyphenation pattern has been}%
\typeout{** loaded for the language `#1'. Using the pattern for}%
\typeout{** the default language instead.}%
\else
\language=\csname l@#1\endcsname
\fi
#2}}

\bibitem{zhao2023learning}
T.~Z. Zhao, V.~Kumar, S.~Levine, and C.~Finn, ``Learning fine-grained bimanual manipulation with low-cost hardware,'' in \emph{RSS}, 2023.

\bibitem{fu2024mobile}
Z.~Fu, T.~Z. Zhao, and C.~Finn, ``Mobile {ALOHA}: Learning bimanual mobile manipulation with low-cost whole-body teleoperation,'' \emph{arXiv:2401.02117}, 2024.

\bibitem{chi2024universal}
C.~Chi, Z.~Xu, C.~Pan, E.~Cousineau, B.~Burchfiel, S.~Feng, R.~Tedrake, and S.~Song, ``Universal manipulation interface: In-the-wild robot teaching without in-the-wild robots,'' \emph{arXiv:2402.10329}, 2024.

\bibitem{allegro}
\url{https://www.allegrohand.com/}.

\bibitem{schunk}
\url{https://schunk.com/us/en/gripping-systems/special-gripper/svh/c/PGR_3161}.

\bibitem{shadow}
\url{https://www.shadowrobot.com/dexterous-hand-series/}.

\bibitem{bicchi2000robotic}
A.~Bicchi and V.~Kumar, ``Robotic grasping and contact: A review,'' in \emph{ICRA}, 2000.

\bibitem{Caccavale2008SixDOFIC}
F.~Caccavale, P.~Chiacchio, A.~Marino, and L.~Villani, ``Six-{DOF} impedance control of dual-arm cooperative manipulators,'' \emph{Transactions on Mechatronics}, 2008.

\bibitem{Sarkar1993DynamicCO}
N.~Sarkar, X.~Yun, and V.~R. Kumar, ``Dynamic control of 3-{D} rolling contacts in two-arm manipulation,'' in \emph{ICRA}, 1993.

\bibitem{platt2004manipulation}
R.~Platt, A.~H. Fagg, and R.~A. Grupen, ``{Manipulation Gaits: Sequences of Grasp Control Tasks},'' in \emph{ICRA}, 2004.

\bibitem{ott2006humanoid}
C.~Ott, O.~Eiberger, W.~Friedl, B.~Bauml, U.~Hillenbrand, C.~Borst, A.~Albu-Schaffer, B.~Brunner, H.~Hirschmuller, S.~Kielhofer, R.~Konietschke, M.~Suppa, T.~Wimbock, F.~Zacharias, and G.~Hirzinger, ``A humanoid two-arm system for dexterous manipulation,'' in \emph{Humanoids}, 2006.

\bibitem{steffen2010bimanual}
J.~Steffen, C.~Elbrechter, R.~Haschke, and H.~Ritter, ``Bio-inspired motion strategies for a bimanual manipulation task,'' in \emph{Humanoids}, 2010.

\bibitem{vahrenkamp2011bimanual}
N.~Vahrenkamp, M.~Przybylski, T.~Asfour, and R.~Dillmann, ``Bimanual grasp planning,'' in \emph{Humanoids}, 2011.

\bibitem{chen2022bidex}
Y.~Chen, T.~Wu, S.~Wang, X.~Feng, J.~Jiang, Z.~Lu, S.~McAleer, H.~Dong, S.-C. Zhu, and Y.~Yang, ``Towards human-level bimanual dexterous manipulation with reinforcement learning,'' in \emph{NeurIPS}, 2022.

\bibitem{zakka2023robopianist}
K.~Zakka, P.~Wu, L.~Smith, N.~Gileadi, T.~Howell, X.~B. Peng, S.~Singh, Y.~Tassa, P.~Florence, A.~Zeng, and P.~Abbeel, ``Robo{P}ianist: Dexterous piano playing with deep reinforcement learning,'' in \emph{CoRL}, 2023.

\bibitem{qi2022hand}
H.~Qi, A.~Kumar, R.~Calandra, Y.~Ma, and J.~Malik, ``In-hand object rotation via rapid motor adaptation,'' in \emph{CoRL}, 2022.

\bibitem{chen2022visual}
T.~Chen, M.~Tippur, S.~Wu, V.~Kumar, E.~Adelson, and P.~Agrawal, ``Visual dexterity: In-hand dexterous manipulation from depth,'' \emph{Science Robotics}, 2023.

\bibitem{openai2018learning}
OpenAI, M.~Andrychowicz, B.~Baker, M.~Chociej, R.~Józefowicz, B.~McGrew, J.~Pachocki, A.~Petron, M.~Plappert, G.~Powell, A.~Ray, J.~Schneider, S.~Sidor, J.~Tobin, P.~Welinder, L.~Weng, and W.~Zaremba, ``Learning dexterous in-hand manipulation,'' \emph{IJRR}, 2019.

\bibitem{handa2023dextreme}
A.~Handa, A.~Allshire, V.~Makoviychuk, A.~Petrenko, R.~Singh, J.~Liu, D.~Makoviichuk, K.~Van~Wyk, A.~Zhurkevich, B.~Sundaralingam, Y.~Narang, J.-F. Lafleche, D.~Fox, and G.~State, ``Dextreme: Transfer of agile in-hand manipulation from simulation to reality,'' in \emph{ICRA}, 2023.

\bibitem{huang2023dynamic}
B.~Huang, Y.~Chen, T.~Wang, Y.~Qin, Y.~Yang, N.~Atanasov, and X.~Wang, ``Dynamic handover: Throw and catch with bimanual hands,'' in \emph{CoRL}, 2023.

\bibitem{lin2024twisting}
T.~Lin, Z.-H. Yin, H.~Qi, P.~Abbeel, and J.~Malik, ``Twisting lids off with two hands,'' \emph{arXiv:2403.02338}, 2024.

\bibitem{billard2008survey}
A.~Billard, S.~Calinon, R.~Dillmann, and S.~Schaal, ``Survey: Robot programming by demonstration,'' \emph{Springer handbook of robotics}, 2008.

\bibitem{hussein2017imitation}
A.~Hussein, M.~M. Gaber, E.~Elyan, and C.~Jayne, ``Imitation learning: A survey of learning methods,'' \emph{ACM Computing Surveys}, 2017.

\bibitem{ravichandar2020recent}
H.~Ravichandar, A.~S. Polydoros, S.~Chernova, and A.~Billard, ``Recent advances in robot learning from demonstration,'' \emph{Annual review of control, robotics, and autonomous systems}, 2020.

\bibitem{grannen2023stabilize}
J.~Grannen, Y.~Wu, B.~Vu, and D.~Sadigh, ``Stabilize to {A}ct: Learning to coordinate for bimanual manipulation,'' in \emph{CoRL}, 2023.

\bibitem{wang2024dexcap}
C.~Wang, H.~Shi, W.~Wang, R.~Zhang, L.~Fei-Fei, and C.~K. Liu, ``Dex{C}ap: Scalable and portable mocap data collection system for dexterous manipulation,'' \emph{arXiv:2403.07788}, 2024.

\bibitem{fang2023low}
H.~Fang, H.-S. Fang, Y.~Wang, J.~Ren, J.~Chen, R.~Zhang, W.~Wang, and C.~Lu, ``Low-cost exoskeletons for learning whole-arm manipulation in the wild,'' in \emph{ICRA}, 2023.

\bibitem{seo2023deep}
M.~Seo, S.~Han, K.~Sim, S.~H. Bang, C.~Gonzalez, L.~Sentis, and Y.~Zhu, ``Deep imitation learning for humanoid loco-manipulation through human teleoperation,'' in \emph{Humanoids}, 2023.

\bibitem{wu2023gello}
P.~Wu, Y.~Shentu, Z.~Yi, X.~Lin, and P.~Abbeel, ``Gello: A general, low-cost, and intuitive teleoperation framework for robot manipulators,'' \emph{arXiv:2309.13037}, 2023.

\bibitem{arunachalam2023dexterous}
S.~P. Arunachalam, S.~Silwal, B.~Evans, and L.~Pinto, ``Dexterous imitation made easy: A learning-based framework for efficient dexterous manipulation,'' in \emph{ICRA}, 2023.

\bibitem{arunachalam2023holo}
S.~P. Arunachalam, I.~G{\"u}zey, S.~Chintala, and L.~Pinto, ``Holo-{D}ex: Teaching dexterity with immersive mixed reality,'' in \emph{ICRA}, 2023.

\bibitem{qin2022one}
Y.~Qin, H.~Su, and X.~Wang, ``From one hand to multiple hands: Imitation learning for dexterous manipulation from single-camera teleoperation,'' \emph{RA-L}, 2022.

\bibitem{sivakumar2022robotic}
A.~Sivakumar, K.~Shaw, and D.~Pathak, ``Robotic telekinesis: Learning a robotic hand imitator by watching humans on youtube,'' in \emph{RSS}, 2022.

\bibitem{iyer2024open}
A.~Iyer, Z.~Peng, Y.~Dai, I.~Guzey, S.~Haldar, S.~Chintala, and L.~Pinto, ``Open teach: A versatile teleoperation system for robotic manipulation,'' \emph{arXiv:2403.07870}, 2024.

\bibitem{Calandra2018More}
R.~Calandra, A.~Owens, D.~Jayaraman, W.~Yuan, J.~Lin, J.~Malik, E.~H. Adelson, and S.~Levine, ``More than a feeling: Learning to grasp and regrasp using vision and touch,'' \emph{RA-L}, 2018.

\bibitem{qi2023general}
H.~Qi, B.~Yi, S.~Suresh, M.~Lambeta, Y.~Ma, R.~Calandra, and J.~Malik, ``General in-hand object rotation with vision and touch,'' in \emph{CoRL}, 2023.

\bibitem{smith2021active}
E.~Smith, D.~Meger, L.~Pineda, R.~Calandra, J.~Malik, A.~Romero~Soriano, and M.~Drozdzal, ``Active 3d shape reconstruction from vision and touch,'' in \emph{NeurIPS}, 2021.

\bibitem{suresh2023neural}
S.~Suresh, H.~Qi, T.~Wu, T.~Fan, L.~Pineda, M.~Lambeta, J.~Malik, M.~Kalakrishnan, R.~Calandra, M.~Kaess, J.~Ortiz, and M.~Mukadam, ``Neural feels with neural fields: Visuo-tactile perception for in-hand manipulation,'' \emph{arXiv:2312.13469}, 2023.

\bibitem{suresh2022shapemap}
S.~Suresh, Z.~Si, J.~G. Mangelson, W.~Yuan, and M.~Kaess, ``Shapemap 3-d: Efficient shape mapping through dense touch and vision,'' in \emph{ICRA}, 2022.

\bibitem{xu2023tandem3d}
J.~Xu, H.~Lin, S.~Song, and M.~Ciocarlie, ``Tandem3d: Active tactile exploration for 3d object recognition,'' in \emph{ICRA}, 2023.

\bibitem{sunil2023visuotactile}
N.~Sunil, S.~Wang, Y.~She, E.~Adelson, and A.~R. Garcia, ``Visuotactile affordances for cloth manipulation with local control,'' in \emph{CoRL}, 2022.

\bibitem{guzey2023dexterity}
I.~Guzey, B.~Evans, S.~Chintala, and L.~Pinto, ``Dexterity from touch: Self-supervised pre-training of tactile representations with robotic play,'' in \emph{CoRL}, 2023.

\bibitem{guzey2023see}
I.~Guzey, Y.~Dai, B.~Evans, S.~Chintala, and L.~Pinto, ``See to touch: Learning tactile dexterity through visual incentives,'' \emph{arXiv:2309.12300}, 2023.

\bibitem{akhtar2021ability}
A.~Akhtar, J.~A. Austin, J.~M. Cornman, D.~M. Bala, and Z.~Wang, ``System and method for an advanced prosthetic hand,'' Mar 2021.

\bibitem{oculus_reader}
\url{https://github.com/rail-berkeley/oculus_reader}.

\bibitem{feix2015grasp}
T.~Feix, J.~Romero, H.-B. Schmiedmayer, A.~M. Dollar, and D.~Kragic, ``The grasp taxonomy of human grasp types,'' \emph{Transactions on human-machine systems}, 2015.

\bibitem{chi2023diffusion}
C.~Chi, S.~Feng, Y.~Du, Z.~Xu, E.~Cousineau, B.~Burchfiel, and S.~Song, ``Diffusion policy: Visuomotor policy learning via action diffusion,'' in \emph{RSS}, 2023.

\bibitem{ho2020denoising}
J.~Ho, A.~Jain, and P.~Abbeel, ``Denoising diffusion probabilistic models,'' in \emph{NeurIPS}, 2020.

\bibitem{he2016deep}
K.~He, X.~Zhang, S.~Ren, and J.~Sun, ``Deep residual learning for image recognition,'' in \emph{CVPR}, 2016.

\bibitem{ioffe2015batch}
S.~Ioffe and C.~Szegedy, ``Batch normalization: Accelerating deep network training by reducing internal covariate shift,'' in \emph{ICML}, 2015.

\bibitem{wu2018group}
Y.~Wu and K.~He, ``Group normalization,'' in \emph{ECCV}, 2018.

\bibitem{kingma2014adam}
D.~P. Kingma and J.~Ba, ``Adam: A method for stochastic optimization,'' in \emph{ICLR}, 2015.

\bibitem{loshchilov2017decoupled}
I.~Loshchilov and F.~Hutter, ``Decoupled weight decay regularization,'' \emph{arXiv preprint arXiv:1711.05101}, 2017.

\bibitem{li2023efficient}
Y.~Li, C.~Pan, H.~Xu, X.~Wang, and Y.~Wu, ``Efficient bimanual handover and rearrangement via symmetry-aware actor-critic learning,'' in \emph{ICRA}, 2023.

\bibitem{song2021maximum}
Y.~Song, C.~Durkan, I.~Murray, and S.~Ermon, ``Maximum likelihood training of score-based diffusion models,'' in \emph{NeurIPS}, 2021.

\end{thebibliography}
\secv

\end{document}